\documentclass[11pt,a4paper]{article}

\usepackage[ngerman,english]{babel}
\usepackage{amsxtra,amstext,amssymb,amsthm}
\usepackage{booktabs}
\usepackage{nicefrac}
\usepackage{xspace}
\usepackage[noadjust]{cite}
\usepackage{url}
\usepackage{graphics,color}
\usepackage[colorlinks]{hyperref}
\definecolor{linkblue}{rgb}{0.1,0.1,0.8}
\hypersetup{colorlinks=true,linkcolor=black,filecolor=black,urlcolor=black,citecolor=black}
\usepackage[algo2e,ruled,vlined,linesnumbered]{algorithm2e}
\SetKwFor{Function}{function}{is}{end function}
\newcommand{\assign}{\leftarrow}
\SetKw{Input}{input}
\SetKw{Output}{output}
\SetKw{Initialization}{initialization}
\SetAlgoCaptionLayout{small}

\newtheorem{theorem}{Theorem}
\newtheorem{lemma}[theorem]{Lemma}
\newtheorem{proposition}[theorem]{Proposition}
\newtheorem{corollary}[theorem]{Corollary}
\newtheorem{definition}[theorem]{Definition}
\newtheorem{algorithm}[theorem]{Algorithm}

\newcommand{\ignore}[1]{}

\allowdisplaybreaks[1]

\newcommand{\N}{\mathbb{N}}
\newcommand{\R}{\mathbb{R}}
\renewcommand{\epsilon}{\varepsilon}

\newcommand{\F}{\mathcal{F}}

\newcommand{\switchIfDistanceOne}{{\tt switchIfDistanceOne}\xspace}
\newcommand{\randomWhereDifferent}{{\tt randomWhereDifferent}\xspace}
\newcommand{\uniformSample}{{\tt uniformSample}\xspace}

\renewcommand{\complement}{{\tt complement}\xspace}
\newcommand{\flipOneWhereDifferent}{{\tt flipOneWhereDifferent}\xspace}
\newcommand{\flipKWhereDifferent}{{\tt flipKWhereDifferent}\xspace}

\newcommand{\chooseConsistent}{{\tt chooseConsistent}\xspace}
\newcommand{\chooseConsistentSub}{{\tt chooseConsistentSelected}\xspace}

\newcommand{\chooseConsistentSubSimple}{{\tt chooseConsistentSelected}_{f(x^1\sigma)-f_\sigma,\ldots,f(x^r\sigma)-f_\sigma}(x^1\sigma,\ldots,x^r\sigma,\overline{\alpha}\sigma,\alpha\sigma)\xspace}
\newcommand{\update}{{\tt update}\xspace}
\newcommand{\optimizeSubset}{{\tt optimizeSelected}\xspace}

\newcommand{\onemax}{\textsc{OneMax}\xspace}
\newcommand{\om}{\textsc{Om}\xspace}
\newcommand{\leadingones}{\textsc{LeadingOnes}\xspace}
\newcommand{\lo}{\textsc{Lo}\xspace}

\newcommand{\Monotone}{\textsc{Monotone}\xspace}

\DeclareMathOperator{\id}{id}
\begin{document}
\title{Faster Black-Box Algorithms Through Higher Arity Operators}

\author{}
\date{}

\author{Benjamin Doerr\\Max-Planck-Institut f\"ur Informatik\\Campus E1 4\\66123 Saarbr\"ucken, Germany
\and Timo K\"otzing\\Max-Planck-Institut f\"ur Informatik\\Campus E1 4\\66123 Saarbr\"ucken, Germany\thanks{Timo K{\"o}tzing was supported by the Deutsche Forschungsgemeinschaft (DFG) under grant  NE~1182/5-1.}
\and Daniel Johannsen\\Max-Planck-Institut f\"ur Informatik\\Campus E1 4\\66123 Saarbr\"ucken, Germany
\and Per Kristian Lehre\\DTU Informatics\\Technical University of Denmark\\2800 Lyngby, Denmark\thanks{Supported by Deutsche Forschungsgemeinschaft (DFG) 
under grant no. WI~3552/1-1.}
\and Markus Wagner\\Max-Planck-Institut f\"ur Informatik\\Campus E1 4\\66123 Saarbr\"ucken, Germany
\and Carola Winzen\\Max-Planck-Institut f\"ur Informatik\\Campus E1 4\\66123 Saarbr\"ucken, Germany\thanks{Carola Winzen is a recipient of the Google Europe Fellowship in Randomized Algorithms, and this research is supported in part by this Google Fellowship.}}

\maketitle 
%
%

{\sloppy
\begin{abstract}
We extend the work of Lehre and Witt (GECCO 2010) on the unbiased
black-box model by considering higher arity variation operators. In
particular, we show that already for binary operators the black-box
complexity of \leadingones drops from $\Theta(n^2)$ for unary operators to $O(n \log n)$. For \onemax, the $\Omega(n \log n)$ unary black-box complexity drops to $O(n)$ in the binary case. For $k$-ary operators, $k \leq n$, the \onemax-complexity further decreases to $O(n/\log k)$. 

\end{abstract}


\section{Introduction}%
\label{sec:introduction}%

\begin{table*}[b]
\begin{center}
\begin{tabular}{@{}ll|ll|ll@{}}\toprule
Model& Arity & \onemax & &\multicolumn{2}{l}{\leadingones}\\
\midrule
unbiased & $1$ & $\Theta(n \log n)$ & \cite{LehreW10}& $\Theta(n^2)$ &\cite{LehreW10}\\
unbiased & $1<k \leq n$ & \textbf{$O(n /\log k)$} &(here)&
\textbf{$O(n \log n)$} & (here)\\
\hline
unrestricted & n/a & $\Omega(n /\log n)$ & \cite{DrosteJW06} & $\Omega(n)$ &\cite{DrosteJW06}\\
&& $O(n /\log n)$ & \cite{Anil2009BlackBox}& &\\\bottomrule
\end{tabular}
\end{center}
\caption{Black-Box Complexity of \onemax and \leadingones. Note that upper bounds for the unbiased unary black-box complexity immediately carry over to higher arities. Similarly, lower bounds for the unrestricted black-box model also hold for the unbiased model.}
\label{table}
\end{table*}

When we analyze the optimization time of randomized search heuristics, we typically assume that the heuristic does not know anything about the objective function apart from its membership in a large class of functions, e.g., linear or monotone functions. 
Thus, the function is typically considered to be given as a black-box, i.e., in order to optimize the function, the algorithm needs to query the function values of various search points. 
The algorithm may then use the information on the function values to create new search points. 
We call the minimum number of function evaluations needed for a randomized search heuristic to optimize any function $f$ of a given function class $\F$ the \emph{black-box complexity} of $\F$.
We may restrict the algorithms with respect to how it creates new search points from the information collected in previous steps. 
Intuitively, the stronger restrictions that are imposed on which search points the algorithm can query next, the larger the black-box complexity of the function class.

Black-box complexity for search heuristics was introduced in 2006 by Droste, Jansen, and Wegener~\cite{DrosteJW06}. 
We call their model the \emph{unrestricted} black-box model as it imposes few restrictions on how the algorithm may create new search points from the information at hand. 
This model was the first attempt towards creating a complexity theory for randomized search heuristics. 
However, the authors prove bounds that deviate from those known for well-studied search heuristics, such as random local search and evolutionary algorithms. 
For example, the well-studied function class \onemax has an unrestricted black-box complexity of $\Theta(n/\log n)$ whereas standard search heuristics only achieve a $\Omega(n \log n)$ runtime. 
Similarly, the class \leadingones has a linear unrestricted black-box complexity but we typically observe a $\Omega(n^2)$ behavior for standard heuristics.

These gaps, among other reasons, motivated Lehre and Witt~\cite{LehreW10} to propose an alternative model. 
In their \emph{unbiased} black-box model the algorithm may only invoke a so-called unbiased variation operator to create new search points. 
A variation operator returns a new search point given one or more previous search points.
Now, intuitively, the unbiasedness condition implies that the variation operator is symmetric with respect to the bit values and bit positions. 
Or, to be more precise, it must be invariant under Hamming-automorphisms. 
We give a formal definition of the two black-box models in Section~\ref{sec:preliminaries}. 

Among other problem instances, Lehre and Witt analyze the unbiased black-box complexity of the two function classes \onemax and \leadingones. 
They can show that the complexity of \onemax and \leadingones match the above mentioned $\Theta(n \log n)$ and, respectively, $\Theta(n^2)$ bounds, if we only allow \emph{unary} operators. I.e., if the variation operator may only use the information from at most one previously queried search point, the unbiased black-box complexity matches the runtime of the well-known $(1+1)$ Evolutionary Algorithm. 

In their first work, Lehre and Witt give no results on the black-box complexity of \emph{higher arity} models. A variation operator is said to be of arity $k$ if it creates new search points by recombining up to $k$ previously queried search points. 
We are interested in higher arity black-box models because they include commonly used search heuristics which are not covered by the unary operators. Among such heuristics are evolutionary algorithms that employ uniform crossover, particle swarm optimization~\cite{KESwarmIntelligence}, ant colony optimization~\cite{Dorigo2004ACO} and estimation of distribution algorithms~\cite{Larranaga2002EDA}. 

Although search heuristics that employ higher arity operators are poorly understood from a theoretical point of view, there are some results proving that situations exist where higher arity is helpful. For example, Doerr, Klein, and Happ~\cite{Doerr2008Crossover} show that a concatenation operator reduces the runtime on the all-pairs shortest path problem. Refer to the same paper for further references. 

Extending the work from Lehre and Witt, we analyze higher arity black-box complexities of \onemax and \leadingones. 
In particular, we show that, surprisingly, the unbiased black-box complexity drops from $\Theta(n^2)$ in the unary case to $O(n\log n)$ for \leadingones and from $\Theta(n\log n)$ to an at most linear complexity for \onemax. 
As the bounds for unbiased unary black-box complexities immediately carry over to all higher arity unbiased black-box complexities, we see that increasing the arity of the variation operators provably helps to decrease the complexity. 
We are optimistic that the ideas developed to prove the bounds can be further exploited to achieve reduced black-box complexities also for other function classes.

In this work, we also prove that increasing the arity further does again help. In particular, we show that for every $k\leq n$, the unbiased $k$-ary black-box complexity of \onemax can be bounded by $O(n/\log k)$. 
This bound is optimal for $k=n$, because the unbiased black-box
complexity can always be bounded below by the unrestricted black-box
complexity, which is known to be $\Omega(n/\log n)$ for \onemax ~\cite{DrosteJW06}.

Note that a comparison between the unrestricted black-box complexity and the unbiased black-box complexity of \leadingones cannot be derived that easily. The asymptotic linear unrestricted black-box complexity mentioned above is only known to hold for a subclass of the class \leadingones considered in this work.

Table~\ref{table} summarizes the results obtained in this paper, and provides a comparison with known results on black-box complexity of \onemax and \leadingones.
  
\section{Unrestricted and Unbiased Black-Box Complexities}%
\label{sec:preliminaries}%

In this section, we formally define the two black-box models by Droste, Jansen, and Wegener~\cite{DrosteJW06}, and Lehre and Witt~\cite{LehreW10}. We call the first model the \emph{unrestricted black-box model}, and the second model the \emph{unbiased black-box model}. Each model specifies a class of algorithms. The black-box complexity of a function class is then defined with respect to the algorithms specified by the corresponding model. We start by describing the two models, then provide the corresponding definitions of black-box complexity.

In both models, one is faced with a class of pseudo-Boolean functions~$\mathcal{F}$ that is known to the algorithm. An adversary chooses a function~$f$ from this class. The function~$f$ itself remains unknown to the algorithm. The algorithm can only gain knowledge about the function~$f$ by querying an oracle for the function value of search points. The goal of the algorithm is to find a globally optimal search point for the function. Without loss of generality, we consider maximization as objective. 
The two models differ in the information available to the algorithm, and the search points that can be queried.

Let us begin with some notation. Throughout this paper, we consider
the maximization of pseudo-Boolean functions $f:\{0,1\}^n \rightarrow
\R$. In particular,~$n$ will always denote the length of the bitstring
to be optimized. For a bitstring $x \in \{0,1\}^n$, we write
$x=x_1\cdots x_n$. For convenience, we denote the positive integers by
$\N$. For $k\in \N$, we use the notion $[k]$ as a shorthand for the
set $\{1,\ldots, k\}$. Analogously, we define $[0..k]:=[k] \cup
\{0\}$. Furthermore, let~$S_k$ denote the set of all permutations of
$[k]$. With slight abuse of notation, we write
$\sigma(x):=x_{\sigma(1)}\cdots x_{\sigma(n)}$ for~$\sigma \in
S_n$. Furthermore, the bitwise exclusive-or is denoted by
$\oplus$. For any bitstring $x$ we denote its complement
by~$\overline{x}$. Finally, we use standard notation for
asymptotic growth of functions (see, e.g., \cite{C01Algorithms}).
In particular, we denote by~$o_n(g(n))$ the set of all functions 
$f$ that satisfy $\lim_{n\rightarrow\infty} f(n)/g(n)=0.$


The \emph{unrestricted black-box model} contains all algorithms which can be formalized as in
Algorithm~\ref{alg:unrestrictedAlgo}. A basic feature is that this scheme does not force any relationship between the search points of subsequent queries. Thus, this model contains a broad class of algorithms. 



\begin{algorithm2e}
 Choose a probability distribution $p_0$ on $\{0, 1\}^n$.\\
 Sample $x^0$ according to $p_0$ and query $f(x^0)$.\\
 \For{$t=1,2,3,\ldots$ until termination condition met}{
  Depending on $((x^0,f(x^0)), \ldots, (x^{t-1},
  f(x^{t-1})))$, choose\\
  $\quad$ a probability distribution $p^{t}$ on $\{0,
  1\}^n$. \\
  Sample $x^t$ according to $p^t$, and query $f(x^t)$.
 }
\caption{Unrestricted Black-Box Algorithm}
\label{alg:unrestrictedAlgo}
\end{algorithm2e}






To exclude some algorithms whose behavior does not resemble those of typical
search-heuristics, one can impose further restrictions.
The \emph{unbiased black-box model} introduced in~\cite{LehreW10}
restricts Algorithm~\ref{alg:unrestrictedAlgo} in two ways. First, the
decisions made by the algorithm only depends on the observed fitness
values, and not the actual search points. Second, the algorithm can
only query search points obtained by variation operators that are
unbiased in the following sense. By imposing these two restrictions,
the black-box complexity matches the runtime of popular search
heuristics on example functions.




\begin{definition} \textup{\textsc{(Unbiased $k$-ary variation operator~\cite{LehreW10})}}\label{def:unbiased-variation}
 Let $k \in \N$. An \emph{unbiased $k$-ary distribution} $D(\cdot\mid x^1,\ldots,x^k)$ is a conditional probability distribution over $\{0,1\}^n$, such that for all bitstrings $y,z \in \{0,1\}^n$, and each permutation $\sigma \in S_n$, the following two conditions hold.

 \begin{enumerate}
 \item[(i)] $D(y \mid x^1, \ldots, x^k) = D(y \oplus z \mid x^1 \oplus z, \ldots, x^k \oplus z)$,
 \item[(ii)] $D(y \mid x^1, \ldots, x^k) = D(\sigma(y) \mid \sigma(x^1), \ldots, \sigma(x^k))$\,.
 \end{enumerate}

 An \emph{unbiased $k$-ary variation operator} $p$ is a $k$-ary operator which samples its output according to an unbiased $k$-ary distribution.
\end{definition}

The first condition in Definition~\ref{def:unbiased-variation} is
referred to as $\oplus$-\emph{invariance}, and the second condition is
referred to as \emph{permutation invariance}. Note that the
combination of these two conditions can be characterized as invariance
under Hamming-automorphisms: $D(\cdot \mid x^1, \ldots,
x^k)$ is unbiased if and only if, for all $\alpha : \{0,1\}^n \rightarrow
\{0,1\}^n$ preserving the Hamming distance and all bitstrings
$y$,
$D(y \mid x^1, \ldots, x^k) =
D(\alpha(y) \mid \alpha(x^1), \ldots, \alpha(x^k))$. We refer to
$1$-ary and $2$-ary variation operators as \emph{unary} and
\emph{binary} variation operators, respectively. The unbiased $k$-ary black-box model contains all algorithms which follow the scheme of Algorithm~\ref{alg:unbiasedAlgo}. While being a restriction of the old model, the unbiased model still captures the most widely studied search heuristics, including most evolutionary algorithms, simulated annealing and random local search.

Note that in line 5 of Algorithm~\ref{alg:unbiasedAlgo}, $y^1, \ldots, y^k$ don't necessarily have to be the $k$ \emph{immediately previously} queried ones. That is, the algorithm is allowed to choose any $k$ previously sampled search points.
 
We now define black-box complexity formally. We will use query complexity as the cost model, where the algorithm is only charged for queries to the oracle, and all other computation is free. The runtime $T_{A,f}$ of a randomized algorithm $A$ on a function $f\in\mathcal{F}$ is hence the expected number of oracle queries until the optimal search point is queried for the first time. The expectation is taken with respect to the random choices made by the
algorithm.

\begin{definition}[Black-box complexity]
 The complexity of a class of pseudo-Boolean functions
 $\mathcal{F}$ with respect to a class of algorithms $\mathcal{A}$,
 is defined as 
 $T_{\mathcal{A}}(\mathcal{F}):=\min_{A\in\mathcal{A}}\max_{f\in\mathcal{F}} T_{A,f}.$
\end{definition}

The \emph{unrestricted black-box complexity} is the 
complexity with respect to the algorithms covered by
Algorithm~\ref{alg:unrestrictedAlgo}. For any given $k\in\N$, the
\emph{unbiased $k$-ary black-box complexity} is the complexity with respect to the algorithms covered by
Algorithm~\ref{alg:unbiasedAlgo}. Furthermore, the
\emph{unbiased $\ast$-ary black-box complexity} is the complexity with respect to the algorithms covered by
Algorithm~\ref{alg:unbiasedAlgo}, without limitation on the arity of the operators used.

It is easy to see that every
unbiased $k$-ary operator $p$ can be simulated by an unbiased
$(k{+}1)$-ary operator $p'$ defined as $p'(z\mid
x^1,\ldots,x^k,x^{k+1}):=p(z\mid x^1,\ldots,x^{k})$. Hence, the unbiased $k$-ary black-box complexity is an upper bound for the 
unbiased $(k{+}1)$-ary black-box complexity. Similarly, the set of
unbiased black-box algorithms for any arity is contained in the set of
unrestricted black-box algorithms. Therefore, the unrestricted black-box complexity is a lower bound for the unbiased $k$-ary black-box complexity (for all $k \in \N$).

\begin{algorithm2e}
	Sample $x^0$ uniformly at random from $\{0,1\}^n$ and query $f(x^0).$\\
	\For{$t=1,2,3,\ldots$ until termination condition met}{
		Depending on $(f(x^0),\ldots, f(x^{t-1}))$, choose\\
                $\quad$ an unbiased $k$-ary variation operator $p^t$,
                and\\
                $\quad$ $k$ previously queried search points $y^1, \ldots, y^k $.\\
		Sample $x^t$ according to $p^t(y^1,\ldots,y^k)$, and query $f(x^t)$.
}
\caption{Unbiased $k$-ary Black-Box Algorithm}
\label{alg:unbiasedAlgo}
\end{algorithm2e}



\section{The Unbiased $\ast$-Ary Black-Box Complexity of OneMax}%
\label{sec:unbiasedOnemax}%
In this section, we show that the unbiased black-box complexity of $\onemax$ is~$\Theta(n/\log n)$ with a leading constant between one and two.
We begin with the formal definition of the function class $\onemax_n$.
We will omit the subscript ``$n$'' if the size of the input is clear from the context.
\begin{definition}[\onemax]
\label{def:onemax}
For all $n \in \N$ and each $z \in \{0,1\}^n$ we define $\om_z:
\{0,1\}^n \rightarrow \N, x \mapsto |\{ j \in [n] \mid x_j = z_j
\}|$.\footnote{Intuitively, $\om_z$ is the function of $n$ minus the Hamming distance to $z$.} The class $\onemax_n$ is defined as $\onemax_n:= \{\om_z
\mid z \in \{0,1\}^n \}\,.$
\end{definition}
To motivate the definitions, let us briefly mention that we do not further consider the optimization of specific functions such as $\om_{(1,\ldots,1)}$, since they would have an unrestricted black-box complexity of $1$: The algorithm asking for the bitstring $(1,\ldots,1)$ in the first step easily optimizes the function in just one step. Thus, we need to consider some generalizations of these functions.
For the unrestricted black-box model, we already have a lower bound by Droste Jansen, and Wegener~\cite{DrosteJW06}. For the same model, an algorithm which matches this bound in order of magnitude is given by Anil and Wiegand in~\cite{Anil2009BlackBox}.
\begin{theorem}
\label{thm:unrestrictedOnemax}
The unrestricted black-box complexity of $\onemax_n$ is $\Theta(n/\log n)$. Moreover, the leading constant is at least~1.
\end{theorem}

As already mentioned, the lower bound on the complexity of $\onemax_n$ in the unrestricted black-box model from Theorem~\ref{thm:unrestrictedOnemax} directly carries over to the stricter unbiased black-box model.

\begin{corollary}
\label{thm:unbiasedOnemaxLower}
The unbiased $\ast$-ary black-box complexity of $\onemax_n$ is at least $n/\log n$.
\end{corollary}

Moreover, an upper bound on the complexity of $\onemax$ in the unbiased black-box model can be derived using the same algorithmic approach as given for the unrestricted black-box model (compare~\cite{Anil2009BlackBox} and Theorem~\ref{thm:unrestrictedOnemax}). 
\begin{theorem}
\label{thm:unbiasedOnemaxUpper}
The unbiased $\ast$-ary black-box complexity of $\onemax_n$ is at most 
$(1+o_n(1))\frac{2n}{\log n}$.
\end{theorem}

In return, this theorem also applies to the unrestricted black-box model and refines Theorem~\ref{thm:unrestrictedOnemax} by explicitly bounding the leading constant of the \emph{unrestricted} black-box complexity for \onemax by a factor of two of the lower bound. The result in Theorem~\ref{thm:unbiasedOnemaxUpper} is based on Algorithm~\ref{alg:unbiasedOnemax}. This algorithm makes use of the operator \uniformSample that samples a bitstring uniformly at random, which clearly is an unbiased ($0$-ary) variation operator. Further, it makes use of another family of operators: $\chooseConsistent_{u^1,\ldots,u^t}(x^1,\ldots,x^t)$ chooses a $z \in \{0,1\}^n$ uniformly at random such that, for all $i \in [t]$, $\om_z(x^i) = u^i$ (if there exists one, and any bitstring uniformly at random otherwise). It is easy to see that this is a family of unbiased variation operators.

\begin{algorithm2e}
 \Input Integer~$n\in\mathbb{N}$ and function~$f \in \onemax_n$\;
 \Initialization $t \assign \big\lceil\big(1+\frac{4\log\log n}{\log n}\big)\frac{2n}{\log n}\big\rceil$\;
	\Repeat{$f(w)=n$}{
\ForEach{$i\in[t]$}{
		$x^i \assign$ \uniformSample$()$\;
}
$w \assign \chooseConsistent_{f(x^1),\ldots,f(x^t)}(x^1,\ldots,x^t)$\;
	}
	\Output $w$\;
\caption{Optimizing $\onemax$ with unbiased variation operators.}
\label{alg:unbiasedOnemax}
\end{algorithm2e}

\ignore{
Sampling a random point is a . Given a set of points $X=\{x_1,\dots,x_t\}$, it is an unbiased $t$-ary variation operator to choose uniformly at random a point in $\{0,1\}^n$ that has distance $\onemax_z(x)$ from~$x$ for each~$x\in X$. Thus, all operations performed by Algorithm~\ref{alg:unbiasedOnemax} are unbiased and therefore this algorithm respects the unbiased black-box model. Thus, the expected runtime of Algorithm~\ref{alg:unbiasedOnemax} is an upper bound for the unbiased black-box complexity of \onemax.}

An upper bound of~$(1+o_n(1))2n/\log n$ for the expected runtime of Algorithm~\ref{alg:unbiasedOnemax} follows directly from the following theorem which implies that the number of repetitions of steps~4 to~6 follows a geometric distribution with success probability~$1-o_n(1)$. This proves Theorem~\ref{thm:unbiasedOnemaxUpper}.

\begin{theorem}
\label{thm:sampling}
Let~$n$ be sufficiently large (i.\,e., let~$n\ge N_0$ for some fixed constant~$N_0\in\mathbb{N}$). Let~$z\in\{0,1\}^n$ and let~$X$ be a set of~$t\ge\big(1+\frac{4\log\log n}{\log n}\big)\frac{2n}{\log n}$ samples chosen from~$\{0,1\}^n$ uniformly at random and mutually independent. Then the probability that there exists an element~$y\in\{0,1\}^n$ such that~$y \neq z$ and~$\om_y(x)=\om_z(x)$ for all~$x\in X$ is bounded from above by~$2^{-t/2}$.
\end{theorem}

The previous theorem is a refinement of Theorem~1 in~\cite{Anil2009BlackBox}, and its proof follows the proof of Theorem~1 in~\cite{Anil2009BlackBox}, clarifying some inconsistencies\footnote{For example, in the proof of Lemma~1 in~\cite{Anil2009BlackBox} the following claim is made. Let~$d(n)$ be a monotone increasing sequence that tends to infinity. Then for sufficient large~$n$ the sequence $h_d(n)=(\frac{\pi d(n)}{8})^{1/(2\ln n)}$ is bounded away from~$1$ by a constant~$b>1$. Clearly, this is not the case. For example, for~$d(n)=\log n$, the sequence $h_{\log}(n)$ converges to~$1$.} in that proof.
To show Theorem~\ref{thm:sampling}, we first give a bound on a
combinatorial quantity used later in its proof (compare Lemma~1
in~\cite{Anil2009BlackBox}). 

\begin{proposition}
\label{prop:binomial}
For sufficiently large $n$, 
\begin{align*}
t\ge\left(1+\frac{4\log\log n}{\log n}\right)\frac{2n}{\log n},
\end{align*}
and even~$d\in\{2,\dots,n\}$, it holds that  
\begin{equation}
\label{eq:binomial}
\binom{n}{d}\left(\binom{d}{\nicefrac{d}{2}}2^{-d}\right)^t\le 2^{-3t/4}.
\end{equation}
\end{proposition}
\begin{proof}
By Stirling's formula, we have~$\binom{d}{\nicefrac{d}{2}}\le\left(\frac{\pi d}{2}\right)^{-\nicefrac{1}{2}}2^d$. Therefore,
\begin{equation}
\label{eq:stirling}
\binom{n}{d}\left(\binom{d}{\nicefrac{d}{2}}2^{-d}\right)^t\le\binom{n}{d}\left(\frac{\pi d}{2}\right)^{-\nicefrac{t}{2}}.
\end{equation}
We distinguish two cases. First, we consider the case $2\le d<n/(\log n)^3$. By Stirling's formula, it holds that $\binom{n}{d}\le\left(\frac{en}{d}\right)^d$. Thus, we get from (\ref{eq:stirling}) that
\begin{equation}
\label{eq:casea}
\begin{aligned}
\binom{n}{d}\left(\binom{d}{\nicefrac{d}{2}}2^{-d}\right)^t
& \le \left(\frac{en}{d}\right)^d \left(\frac{\pi d}{2}\right)^{-\nicefrac{t}{2}}\\
& = 2^{\left(\frac{2d}{t}\log\left(\frac{en}{d}\right) -\log\left(\frac{\pi d}{2}\right)\right)\frac{t}{2}}.
\end{aligned}
\end{equation}
We bound~$d$ by its minimal value~$2$ and maximal value~$n/(\log n)^3$, and~$t$ by~$2n/\log n$ to obtain
\[
\frac{2d}{t}\log\frac{en}{d} -\log\frac{\pi d}{2}\le\frac{1}{(\log n)^2}\log\frac{en}{2}-\log\pi.
\]
Since the first term on the right hand side converges to 0 and since
$\log\pi>3/2$, the exponent in (\ref{eq:casea}) can be bounded from above by -3t/4, if $n$
is sufficiently large.
Thus, we obtain inequality~(\ref{eq:binomial}) for $2\le d<n/(\log n)^3$.

Next, we consider the case $n/(\log n)^3\le d\le n$. By the binomial formula, it holds that $\binom{n}{d}\le 2^n$. Thus,
\begin{equation}
\label{eq:caseb}
\binom{n}{d} \left(\frac{\pi d}{2}\right)^{-\nicefrac{t}{2}}
\le 2^n \left(\frac{\pi d}{2}\right)^{-\nicefrac{t}{2}}
=2^{\left(\frac{2n}{t}-\log\frac{\pi d}{2}\right)\frac{t}{2}}.
\end{equation}
We bound~$\pi d/2$ by~$n/(\log n)^3$ and~$t$ by~$\big(1+\frac{4\log\log n}{\log n}\big)\frac{2n}{\log n}$ to obtain
\begin{align*}
\frac{2n}{t}-\log\frac{\pi d}{2}
& \leq \frac{\log n}{1+\frac{4\log\log n}{\log n}}-\log (n/(\log n)^3)\\
& = \frac{\log n  }{1+\frac{4\log\log n}{\log n}}-\log n + 3 \log \log n\\
& = \frac{3 \log \log n +\frac{4\log\log n}{\log n}(- \log n  + 3 \log \log n)}{1+\frac{4\log\log n}{\log n}}\\
& = -\frac{\log n-12\log\log n}{\log n+4\log\log n}\log\log n.
\end{align*}
Again, for sufficiently large~$n$ the right hand side becomes smaller than~$-3/2$. We combine the previous inequality with inequalities~(\ref{eq:stirling}) and~(\ref{eq:caseb}) to show inequality~(\ref{eq:binomial}) for $n/(\log n)^3\le d\le n$.
\end{proof}

With the previous proposition at hand, we finally prove Theorem~\ref{thm:sampling}.
\begin{proof}[Proof of Theorem~\ref{thm:sampling}]
Let~$n$ be sufficiently large,~$z\in\{0,1\}^n$, and~$X$ a set of~$t\ge\big(1+\frac{4\log\log n}{\log n}\big)\frac{2n}{\log n}$ samples chosen from~$\{0,1\}^n$ uniformly at random and mutually independent.


For~$d \in [n]$, let~$A_d:=\{y\in\{0,1\}^n\;\big|\; n-\om_z(y) = d\}$ be the set of all points with Hamming distance~$d$ from~$z$. 
Let $d\in [n]$ and $y \in A_d$. We say the point~$y$ is \emph{consistent} with~$x$ if~$\om_y(x) = \om_z(x)$ holds. Intuitively, this means that $\om_y$ is a possible target function, given the fitness of $x$. It is easy to see that~$y$ is consistent with~$x$ if and only if~$x$ and~$y$ (and therefore~$x$ and~$z$) differ in exactly half of the~$d$ bits that differ between~$y$ and~$z$. Therefore, $y$ is never consistent with~$x$ if~$d$ is odd and the probability that~$y$ is consistent with~$x$ is~$\binom{d}{\nicefrac{d}{2}}2^{-d}$ if~$d$ is even.

Let~$p$ be the probability that there exists a point~$y\in\{0,1\}^n\setminus\{z\}$ such that~$y$ is consistent with all~$x\in X$. Then,

\[
p=\Pr\Big(\bigcup_{y\in\{0,1\}^n\setminus\{z\}} \; \bigcap_{x\in X}\text{``$y$ is consistent with~$x$''}\Big).
\]
Thus, by the union bound, we have 
\[
p\le \sum_{y\in\{0,1\}^n\setminus\{z\}}\Pr\Big(\bigcap_{x\in X}\text{``$y$ is consistent with~$x$''}\Big).
\]
Since, for a fixed~$y$, the events~``$y$ is consistent with~$x$'' are mutually independent for all~$x\in X$, it holds that
\[
p \le \sum_{d=1}^n\sum_{y\in A_d}\prod_{x\in X}\Pr(\text{``$y$ is consistent with~$x$''}).
\]
We substitute the probability that a fixed $y\in\{0,1\}^n$ is consistent with a randomly chosen~$x\in\{0,1\}^n$ as given above. Using~$|A_d|=\binom{n}{d}$, we obtain
\[
p \le \sum_{d\in\{1,\dots,n\}\colon d\text{ even}}\binom{n}{d}\left(\binom{d}{\nicefrac{d}{2}}2^{-d}\right)^t
\]
Finally, we apply Proposition~\ref{prop:binomial} and have~$p\le n 2^{-3t/4}$ which concludes the proof since~$n\le 2^{t/4}$ for sufficiently large~$n$ (as $t$ in $\Omega(n/ \log n)$).
\end{proof}

\section{The Unbiased $k$-Ary Black-Box Complexity of OneMax}%
\label{sec:arity}%

In this section, we show that higher arity indeed enables the construction of faster black-box algorithms. In particular, we show the following result.
\begin{theorem}
\label{thm:onemax}
For every $k \in [n]$ with $k \geq 2$, the unbiased $k$-ary black-box complexity of $\onemax_n$ is at most linear in~$n$. Moreover, it is at most $(1+o_k(1))2n/\log k$.
\end{theorem}
This result is surprising, since in~\cite{LehreW10}, Lehre and Witt prove that the unbiased unary black-box complexity of the class of all functions $f$ with a unique global optimum is $\Omega(n \log n)$. Thus, we gain a factor of $\log n$ when switching from unary to binary variation operators.

%

To prove Theorem~\ref{thm:onemax}, we introduce two different algorithms interesting on their own. 
Both algorithms share the idea to track which bits have already been optimized. That way we can avoid flipping them again in future iterations of the algorithm. 

The first algorithm proves that the unbiased binary black-box complexity of $\onemax_n$ is at most linear in~$n$ if the arity is at least two.
For the general case, with $k\geq3$, we give a different algorithm that provides asymptotically better bounds for~$k$  growing in~$n$. 
We use the idea that the whole bitstring can be divided into smaller substrings, and subsequently those can be independently optimized. We show that this is possible, and together with Theorem~\ref{thm:unbiasedOnemaxUpper}, this yields the above bound for $\onemax_n$ in the $k$-ary case for $k\ge 3$.

\subsection{The Binary Case}\label{sec:2ary}

We begin with the binary case. We use the three unbiased variation operators $\uniformSample$ (as described in Section~\ref{sec:unbiasedOnemax}), $\complement$ and $\flipOneWhereDifferent$ defined as follows. 
The unary operator $\complement(x)$ returns the bitwise complement of $x$. 
The binary operator $\flipOneWhereDifferent(x,y)$ returns a copy of $x$, where one of the bits that differ in $x$ and $y$ is chosen uniformly at random and then flipped.
It is easy to see that $\complement$ and $\flipOneWhereDifferent$ are unbiased variation operators. \vspace{-2mm}


\begin{algorithm2e}
 \Input Integer~$n\in\mathbb{N}$ and function~$f\in\onemax_n$\; 
	\Initialization $x \assign \uniformSample()$\; 
	$y \assign \complement(x)$\;
	\Repeat{$f(x) = n$}{
		Choose $b \in \{0,1\}$ uniformly at random\;
      \eIf{$b=1$}{
       	$x' \assign \flipOneWhereDifferent(x,y)$\;
     	\lIf{$f(x') > f(x)$}{$x \assign x'$\;}
			}{
        $y' \assign \flipOneWhereDifferent(y,x)$\;
			\lIf{$f(y') > f(y)$}{$y \assign y'$\;}}
	}
	\Output $x$\;
\caption{Optimizing \onemax with unbiased binary variation operators.}
\label{alg:2aryalgo}
\end{algorithm2e}

\begin{lemma}\label{lemma:onemax-2-ary}
With exponentially small probability of failure, the optimization time of Algorithm~\ref{alg:2aryalgo} on the class $\onemax_n$ is at most $(1+\epsilon)2n$, for all $\epsilon > 0$. The algorithm only involves binary operators.
\end{lemma}


\begin{proof}
We first prove that the algorithm is correct. Assume that the instance
has optimum $z$, for some $z\in \{0,1\}^n$.
We show that the following invariant is satisfied in the beginning of
every iteration of the main loop (steps 4-12):
for all $i \in \left[n\right]$, if $x_i=y_i$, then
$x_i=z_i$. In other words, the positions where $x$ and $y$ have the same
bit value are optimized.  The invariant clearly holds in the first 
iteration, as $x$ and $y$ differ in all bit positions. 
A bit flip is only accepted if 
the fitness value is strictly higher, an event which occurs with
positive probability.
Hence, if the invariant holds in the current iteration, then
it also holds in the following iteration. By induction, the invariant
property now holds in every iteration of the main loop.

We then analyze the runtime of the algorithm. Let $T$ be the number of
iterations needed until $n$ bit positions have been optimized. Due to
the invariant property, this is the same as the time needed to reduce
the Hamming distance between $x$ and $y$ from $n$ to $0$. An iteration
is successful, i.e., the Hamming distance is reduced by 1, with
probability $1/2$ independently of previous trials. The random variable $T$ is
therefore negative binomially distributed with
parameters $n$ and $1/2$. It can be related to a
binomially distributed random variable $X$ with parameters
$2n(1+\varepsilon)$ and $1/2$ by $\Pr(T\geq
2n(1+\varepsilon))=\Pr(X\leq n)$.  Finally, by applying a Chernoff 
bound with respect to $X$, we obtain $\Pr(T\geq 2n(1+\varepsilon))\leq
\exp(-\varepsilon^2n/2(1+\varepsilon))$.%
%
%
%
\end{proof}

%
It is easy to see that Algorithm~\ref{alg:2aryalgo} yields the same bounds on the class of monotone functions, which is defined as follows. 

\begin{definition}[Monotone functions]
\label{def:monotone}
Let $n \in \N$ and let $z \in \{0,1\}^n$. 
A function $f:\{0,1\}^n \rightarrow \R$ is said to be \emph{monotone with respect to $z$} if for all $y, y'\in \{0,1\}^n$ with $\{i \in [n] \mid y_i = z_i\} \subsetneq \{i \in [n] \mid y'_i = z_i\}$ it holds that $f(y) < f(y')$. The class $\Monotone_n$ contains all such functions that are monotone with respect to some $z \in \{0,1\}^n$.
\end{definition}

Now, let $f$ be a monotone function with respect to $z$ and let $y$ and $y'$ be two bitstrings which differ only in the $i$-th position. Assume that $y_i \neq z_i$ and $y_i' = z_i$. It follows from the monotonicity of $f$ that $f(y) < f(y')$. 
Consequently, Algorithm~\ref{alg:2aryalgo} optimizes $f$ as fast as any function in $\onemax_n$. 

\begin{corollary}
The unbiased binary black-box complexity of $\Monotone_n$ is $O(n)$.
\end{corollary}

Note that $\Monotone_n$ strictly includes the class of linear functions with non-zero weights.

%
\subsection{Proof of Theorem \ref{thm:onemax} for Arity
  $k\geq 3$}
\label{sec:geq3ary}
For the case of arity $k \geq 3$ we analyze the following
Algorithm~\ref{alg:karyalgoShort} and show that its optimization time
on $\onemax_n$ is at most $(1+o_k(1))2n/\log k$.  Informally, the
algorithm splits the bitstring into blocks of length~$k$. The $n/k$
blocks are then optimized separately using a variant of
Algorithm~\ref{alg:unbiasedOnemax}, each in expected time
$(1-o_k(1))2k/\log k$.

In detail, Algorithm~\ref{alg:karyalgoShort} maintains its state
using three bitstrings $x, y$ and $z$. Bitstring $x$ represents the
preliminary solution. The positions in which bitstrings $x$ and $y$
differ represent the remaining blocks to be optimized, and the
positions in which bitstrings $y$ and $z$ differ represent the current
block to be optimized.  Due to permutation invariance, it can be
assumed without loss of generality that the bitstrings can be
expressed by $x=\alpha\overline{\beta}\gamma$,
$y=\overline{\alpha}\beta\gamma$, and~$z=\alpha\beta\gamma$, see
Step~6 of Algorithm~\ref{alg:karyalgoShort}. The algorithm uses
an operator called $\flipKWhereDifferent_\ell$ to select a new block
of size~$\ell$ to optimize.  The selected block is
optimized by calling the subroutine $\optimizeSubset_{n,\ell}$, and
the optimized block is inserted into the preliminary solution using
the operator \update.

The operators in Algorithm~\ref{alg:karyalgoShort} are defined as
follows. The operator $\flipKWhereDifferent_{k}(x,y)$ generates the
bitstring~$z$. This is done by making a copy of~$y$, choosing
$\ell:=\min\{k,H(x,y)\}$ bit positions for which~$x$ and~$y$ differ
uniformly at random, and flipping them. The operator $\update(a,b,c)$
returns a bitstring $a'$ which in each position $i\in[n]$
independently, takes the value $a_i'=b_i$ if $a_i=c_i$, and $a_i'=a_i$
otherwise. Clearly, both these operators are unbiased. The operators
\uniformSample and \complement have been defined in previous sections.

\begin{algorithm2e}[!ht]
\Input 
  Integers~$n,k\in\mathbb{N},$ and function~$f\in\onemax_n$\;
\Initialization 
  $x^1 \assign \uniformSample(), $ 
  $y^1\assign\complement(x),$ and 
  $\tau\assign \lceil\frac{n}{k}\rceil$\; 
\ForEach{$t\in\left[ \tau \right]$}{
  $\ell(t) \assign \min\{k,n-k(t-1)\}$\;
  $z \assign \flipKWhereDifferent_{\ell(t)}(x^t,y^t)$\;
  Assume that $x^t=\alpha\overline{\beta}\gamma,\; y^t=\overline{\alpha}\beta\gamma,$
  and $z=\alpha\beta\gamma$\;
  $w^t\beta\gamma \assign \optimizeSubset_{n,\ell(t)}(\overline{\alpha}\beta\gamma,\alpha\beta\gamma)$\;
  $w^t\overline{\beta}\gamma \assign \update(\alpha\overline{\beta}\gamma,w^t\beta\gamma,\alpha\beta\gamma)$\;
  $x^{t+1}\assign w^t\overline{\beta}\gamma$ and $y^{t+1}\assign w^t\beta\gamma$\;
}
\Output $x^{\tau+1}$\;
\caption{Optimizing \onemax with unbiased $k$-ary variation operators, for $k \geq 3$.}
\label{alg:karyalgoShort}
\end{algorithm2e}

It remains to define the subroutine $\optimizeSubset_{n,k}$. This subroutine is
a variant of Algorithm~\ref{alg:unbiasedOnemax} that only optimizes a
selected block of bit positions positions, and leaves the other blocks
unchanged. The block is represented by the bit positions in which
bitstrings $y$ and $z$ differ. Due to permutation-invariance, we assume
that they are of the form $y=\overline{\alpha}\sigma$ and
$z=\alpha\sigma$, for some bitstrings $\alpha\in\{0,1\}^k$, and
$\sigma\in\{0,1\}^{n-k}$.
The operator $\uniformSample$ in Algorithm~\ref{alg:unbiasedOnemax} is
replaced by a $2$-ary operator defined by: $\randomWhereDifferent(x,y)$
chooses $z$, where for each $i\in[n]$, the value of bit $z_i$ is
$x_i$ or $y_i$ with equal probability. Note that this operator is the
same as the standard uniform crossover operator.
The operator family $\chooseConsistent$ in
Algorithm~\ref{alg:unbiasedOnemax} is replaced by a family of
$(r+2)$-ary operators defined by:
$\chooseConsistentSub_{u^1,\ldots,u^r}(x^1,\ldots, x^r,
\overline{\alpha}\sigma,\alpha\sigma)$ chooses $z\sigma$, where the
prefix $z$ is sampled uniformly at random from the set 
$Z_{u,x}=\{z\in\{0,1\}^k\mid \forall i\in[t]\; \om_{z}(x^i_1x^i_2\cdots
x_k^i)=u^i\}$.  If the set $Z_{u,x}$ is empty, then $z$ is sampled
uniformly at random among all bitstrings of length $k$. 
Informally, the set $Z_{u,x}$ corresponds to the subset of functions
in $\onemax_n$ that are consistent with the function values 
$u^1,u^2,\dots, u^r$ on the inputs $x^1,x^2,\dots, x^r$. It is easy to
see that this operator is unbiased.

\begin{algorithm2e}[!ht]
\Input 
  Integers $n,k\in\mathbb{N},$ and 
  bitstrings $\overline{\alpha}\sigma$ and $\alpha\sigma$, where
  $\alpha\in\{0,1\}^k$ and $\sigma\in\{0,1\}^{n-k}$\;
\Initialization 
  $r \assign \min\Big\{k-2,\big\lceil\big(1+\frac{4\log\log k}{\log k}\big)\frac{2k}{\log k}\big\rceil\Big\}$,
  $\quad f_{\sigma}\assign \frac{f(\alpha\sigma)+f(\overline{\alpha}\sigma)-k}{2}$\;
\Repeat{$f(w\sigma) = k+f_{\sigma}$}{
  \ForEach{$i \in \left[r\right] $}{
    $x^i\sigma \assign \randomWhereDifferent(\alpha\sigma,\overline{\alpha}\sigma)$\;
  }
  $w\sigma \assign \chooseConsistentSubSimple$\;
}
\Output $w\sigma$\;
\caption{$\optimizeSubset$ used in Algorithm~\ref{alg:karyalgoShort}.}
\label{alg:karyalgoSubroutine}
\end{algorithm2e}

\begin{proof}[Proof of Theorem~\ref{thm:onemax} for arity $k\geq3$]
  To prove the \emph{correctness} of the algorithm, assume without
  loss of generality the input $f=\onemax$ for which the correct 
  output is $1^n$.

  We first claim that a call to
  $\optimizeSubset_{n,k}(\overline{\alpha}\sigma,\alpha\sigma)$ will
  terminate after a finite number of iterations with output
  $1^k\sigma$ almost surely. The variable $f_\sigma$ is assigned in
  line 2 of Algorithm \ref{alg:karyalgoSubroutine}, and it is easy to
  see that it takes the value $f_\sigma=f(0^k\sigma)$. It follows from
  linearity of $f$ and from $f(\alpha 0^{n-k})+f(\overline{\alpha}
  0^{n-k})=k$, that $f(w0^{n-k}) = f(w\sigma) - f(0^k\sigma) =
  f(w\sigma) - f_\sigma$. The termination condition
  $f(w\sigma)=k+f_\sigma$ is therefore equivalent to the condition
  $w\sigma=1^k\sigma$. For all $x\in\{0,1\}^k$, it holds that
  $\om_{(1,\ldots,1)}(x)=f(x)$, so $1^k$ is member of the set
  $Z_{u,x}$.  Hence, every invocation of $\chooseConsistentSub$
  returns $1^k\sigma$ with non-zero probability.  Therefore, the
  algorithm terminates after every iteration with non-zero
  probability, and the claim holds.

  We then prove by induction the invariant property that for all
  $t\in[\tau+1]$, and $i\in[n]$, if
  $x_i^t=y_i^t$ then $x_i^t=y_i^t=1$. The invariant clearly holds for
  $t=1$, so assume that the invariant also holds for
  $t=j\leq\tau$. Without loss of generality,
  $x^j=\alpha\overline{\beta}\gamma$,
  $y^j=\overline{\alpha}\beta\gamma$,
  $x^{j+1}=w^j\overline{\beta}\gamma$, and $y^{j+1}=w^j\beta\gamma$.  By
  the claim above and the induction hypothesis, both the common prefix
  $w^j$ and the common suffix $\gamma$ consist of only 1-bits. So
  the invariant holds for $t=j+1$, and by induction also for all
  $t\in[\tau+1]$.

  It is easy to see that for all $t\leq \tau$, the
  Hamming distance between $x^{t+1}=w^t\overline{\beta}\gamma$ and
  $y^{t+1}=w^t\beta\gamma$ is $H(x^{t+1},y^{t+1}) =
  H(\alpha\overline{\beta}\gamma,\overline{\alpha}\beta\gamma)-\ell(t) =
  H(x^t,y^t)-\ell(t)$.  By induction, it therefore holds that
  \begin{align*}
    H(x^{\tau+1},y^{\tau+1}) 
   & = H(x^1,y^1) - \sum_{t=1}^\tau \ell(t) \\
   & = n -\sum_{t=1}^\tau \min\{k,n-k(t-1)\} = 0.
  \end{align*}
  Hence, by the invariant above, the algorithm returns the correct output
  $x^{\tau+1}=y^{\tau+1}=1^n$.

  The \emph{runtime} of the algorithm in each iteration is dominated by the subroutine
  $\optimizeSubset$. 
  Note that by definition, the probability that
  $\chooseConsistentSub_{f(x^1\sigma)-f_\sigma,\ldots,f(x^r\sigma)-f_\sigma}(x^1\sigma,\ldots,
  x^r\sigma, \overline{\alpha}\sigma,\alpha\sigma)$ chooses $z\sigma$ in~$\{0,1\}^n$,
  is the same as the probability that
  $\chooseConsistent_{f(x^1),\ldots,f(x^r)}(x^1,\ldots, x^r)$ chooses
  $z$ in~$\{0,1\}^k$.
  To finish the proof, we distinguish between two cases.

  \emph{Case 1: $k\leq53$.} In this case, it suffices\footnote{
    Assume that the expected runtime is less than $cn$ for some constant
    $c>0$ when $k\leq 53.$ It is necessary to show that 
    $cn\leq 2n/\log k + h(k)2n/\log k$, for some function $h$, 
    where $\lim_{k\rightarrow\infty}
    h(k)\rightarrow 0.$ This can easily be shown by choosing any such 
    function $h,$ where $h(k)\geq c\log k/2$ for $k\leq 53$.} to prove that
  the runtime is $O(n)$. For the case $k=2$, this follows from
  Lemma~\ref{lemma:onemax-2-ary}. For $2<k\leq 53$, it holds that
  $r=k-2$. Each iteration in \optimizeSubset uses $r+1=k-1=O(1)$
  function evaluations, and the probability that \chooseConsistentSub
  optimizes a block of $k$ bits is at least $1-(1-2^{-k})^r=\Omega(1)$
  (when $w=1^k$).  Thus, the expected optimization time for a block is
  $O(1)$, and for the entire bitstring it is at most $(n/k) \cdot O(1)$.

  \emph{Case 2: $k\geq54$.} In this case,
  $r=\big\lceil\big(1+\frac{4\log\log k}{\log k}\big)\frac{2k}{\log
    k}\big\rceil$ holds. 
  Hence, with an analysis analogous to that in the proof of 
  Theorem~\ref{thm:unbiasedOnemaxUpper}, we can show
  that the expected runtime of \optimizeSubset is at most 
  $(1+o_k(1))2(k-2)/\log (k-2)$. Thus, the expected runtime is at 
  most $n/k \cdot (1+o_k(1))2(k-2)/\log (k-2) = (1+o_k(1))2n/\log k$.
\end{proof}

\section{The Complexity of LeadingOnes}
\label{sec:LO}

In this section, we show that allowing $k$-ary variation operators, for $k > 1$, greatly reduces the black-box complexity of the \leadingones functions class, namely from $\Theta(n^2)$ down to $O(n \log n)$.
%
We define the class \leadingones as follows.

\begin{definition}[\leadingones]
\label{def:leadingones}
Let $n \in \N$. Let $\sigma \in S_n$ be a permutation of the set $[n]$
and let $z \in \{0,1\}^n$. The function $\lo_{z,\sigma}$ is
defined via $\lo_{z,\sigma}(x):= \max \{ i \in [0..n] \mid
z_{\sigma(i)} = x_{\sigma(i)}\}$. We set $\leadingones_n:=
\{\lo_{z, \sigma} \mid z \in \{0,1\}^n, \sigma \in S_n \}\,.$
\end{definition}

The class \leadingones is well-studied. Already in 2002, Droste, Jansen and Wegener~\cite{DrosteJW02} proved that the classical $(1+1)$ EA has an expected optimization time of $\Theta(n^2)$ on \leadingones. This bound seems to be optimal among the commonly studied versions of evolutionary algorithms. In~\cite{LehreW10}, the authors prove that the unbiased unary black-box complexity of \leadingones is $\Theta(n^2)$.

Droste, Jansen and Wegener~\cite{DrosteJW06} consider a subclass of $\leadingones_n$, namely $\leadingones_n^0 :=
\{\lo_{z, \id} \mid z \in \{0,1\}^n\}$, where $\id$ denotes the identity mapping on $[n]$. Hence their function class is not permutation invariant. In this restricted setting, they prove a black-box complexity of $\Theta(n)$. Of course, their lower bound of $\Omega(n)$ is a lower bound for the unrestricted black-box complexity of the general $\leadingones_n$ class, and consequently, a lower bound also for the unbiased black-box complexities of this class. 

The following theorem is the main result in this section.

\begin{theorem}
\label{thm:LO}
The unbiased binary black-box complexity of $\leadingones_n$ is $O(n \log n)$.
\end{theorem}

The key ingredient of the two black-box algorithms that yield our upper bound is an emulation of a binary search which determines the (unique) bit that increases the fitness \emph{and} does flip this bit. 
Surprisingly, this can be done already with a binary operator. This works in spite of the fact that we also follow the general approach of the previous section of keeping two individuals $x$ and $y$ such that for all bit positions in which $x$ and $y$ agree, the corresponding bit value equals the one of the optimal solution.

We will use the two unbiased binary variation operators $\randomWhereDifferent$ (as described in Section~\ref{sec:geq3ary}) and $\switchIfDistanceOne$. 
The operator $\switchIfDistanceOne(y,y')$ returns $y'$ if $y$ and $y'$ differ in exactly one bit, and returns $y$ otherwise. It is easy to see that $\switchIfDistanceOne$ is an unbiased variation operators.

We call a pair $(x,y)$ of search points \emph{critical}, if the following two conditions are satisfied. (i) $f(x) \ge f(y)$. (ii) There are exactly $f(y)$ bit-positions $i \in [n]$ such that $x_i = y_i$. The following is a simple observation.

\begin{lemma}\label{lem:losimple}
  Let $f \in \leadingones_n$. If $(x,y)$ is a critical pair, then either $f(x) = n = f(y)$ or $f(x) > f(y)$. 
\end{lemma}

If $f(x) > f(y)$, then the unique bit-position $k$ such that flipping the $k$-th bit in $x$ reduces its fitness to $f(y)$\, -- or equivalently, the unique bit-position such that flipping this bit in $y$ increases $y$'s fitness\, -- shall be called the \emph{critical bit-position}. We also call $f(y)$ the value of the pair $(x,y)$.

Note that the above definition does only use some function values of $f$, but not the particular definition of $f$. If $f = \lo_{\sigma,z}$, then the above implies that $x$ and $y$ are equal on the bit-positions $\sigma(1), \ldots, \sigma(f(y))$ and are different on all other bit-positions. Also, the critical bit-position is $\sigma(f(y)+1)$, and the only way to improve the fitness of $y$ is flipping this particular bit-position (and keeping the positions $\sigma(1), \ldots, \sigma(f(y))$ unchanged). 
The central part of Algorithm~\ref{alg:leadingones}, which is
contained in lines~3 to~9, manages to transform a critical pair of
value $v < n$ into one of value $v+1$ in $O(\log n)$ time. This is analyzed in the following lemma.

\begin{lemma}\label{lem:leadingoneskey}
  Assume that the execution of Algorithm~\ref{alg:leadingones} is before line~4, and that the current value of $(x, y)$ is a critical pair of value $v<n$. 
  Then after an expected number of $O(\log n)$ iterations, the loop in lines~5-9 is left and $(x,y)$ or $(y,x)$ is a critical pair of value $v+1$. 
\end{lemma}

\begin{proof}
  Let $k$ be the critical bit-position of the pair $(x,y)$. Let $y' = x$ be a copy of $x$. Let $J := \{i \in [n] \mid y_i \neq y'_i\}$. Our aim is to flip all bits of $y'$ with index in $J \setminus \{k\}$. 
  
  We define $y''$ by flipping each bit of $y'$ with index in $J$ with probability $1/2$. Equivalently, we can say that $y''_i$ equals $y'_i$ for all $i$ such that $y'_i = y_i$, and is random for all other $i$ (thus, we obtain such $y''$ by applying $\randomWhereDifferent(y,y')$). 
  
  With probability exactly $1/2$, the critical bit was not flipped (``success''), and consequently, $f(y'') > f(y)$. In this case (due to independence), each other bit with index in $J$ has a chance of $1/2$ of being flipped. So with constant probability at least $1/2$, $\{i \in [n] \mid y_i \neq y''_i\} \setminus \{k\}$ is at most half the size of $J \setminus \{k\}$. In this success case, we take $y''$ as new value for $y'$. 
  
  In consequence, the cardinality of $J \setminus \{k\}$ does never increase, and with probability at least $1/4$, it decreases by at least 50\%. Consequently, after an expected number of $O(\log n)$ iterations, we have $|J| = 1$, namely $J = \{k\}$. We check this via an application of $\switchIfDistanceOne$. 
\end{proof}

\vspace{-3mm}
We are now ready to prove the main result of this section.

\begin{proof}[Proof of Theorem~\ref{thm:LO}]
  We regard the following invariant: $(x,y)$ or $(y,x)$ is a critical
  pair. This is clearly satisfied after execution of line~1. From
  Lemma~\ref{lem:leadingoneskey}, we see that a single execution of
  the outer loop does not dissatisfy our invariant. Hence by
  Lemma~\ref{lem:losimple}, our algorithm is correct (provided it
  terminates). The algorithm does indeed terminate, namely in $O(n \log n)$ time, because, again by Lemma~\ref{lem:leadingoneskey}, each iteration of the outer loop increases the value of the critical pair by one. 
\end{proof}  
%

\begin{algorithm2e}
  \Initialization $x \assign \uniformSample()$; $y \assign \complement(x)$\;
	\Repeat{$f(x) = f(y)$}{
  	  \lIf{$f(y) > f(x)$}{$(x,y) \assign (y,x)$\;}
  	  $y' \assign x$\;
  	  \Repeat{$f(y) = f(y')$}{
  	  	$y'' \assign \randomWhereDifferent(y,y')$\;
  	  	\lIf{$f(y'') > f(y)$}{$y' \assign y''$\;}
  	  	$y \assign \switchIfDistanceOne(y,y')$\;
		}
	}
 \Output $x$\;
\caption{Optimizing \leadingones with unbiased binary variation operators.}
\label{alg:leadingones}
\end{algorithm2e}

%
%
%

\vspace{-2mm}
\section{Conclusion and Future Work}%
\label{sec:conclusion}%

We continue the study of the unbiased black-box model introduced in~\cite{LehreW10}. 
For the first time, we analyze variation operators with arity higher than one. Our results show that already two-ary operators can allow significantly faster algorithms.

The problem \onemax cannot be solved in shorter time than $\Omega(n\log n)$ with unary variation operators~\cite{LehreW10}. However, the runtime can be reduced to $O(n)$ with binary operators. The runtime can be decreased even further with higher arities than two. For $k$-ary variation operators, $2\leq k\leq n$, the runtime can be reduced to $O(n/\log k)$, which for $k = n^{\Theta(1)}$ matches the lower bound in the classical black-box model.
%
A similar positive effect of higher arity variation operators can be
observed for the function class \leadingones. While this function
class cannot be optimized faster than $\Omega(n^2)$ with unary
variation operators~\cite{LehreW10}, we show that the runtime can be
reduced to $O(n\log n)$ with binary, or higher arity variation operators.

Despite the restrictions imposed by the unbiasedness conditions,
our analysis demonstrates that black-box algorithms can employ new and more efficient search heuristics with higher arity variation operators. In particular, binary variation operators allow a memory mechanism that can be used to implement binary search on the positions in the bitstring. The algorithm can thereby focus on parts of the bitstring that has not previously been investigated.
%

An important open problem arising from this work is to provide lower bounds in the unbiased black-box model for higher arities than one. Due to the greatly enlarged computational power of black-box algorithms using higher arity operators (as seen in this paper), proving lower bounds in this model seems significantly harder than in the unary model. 

\providecommand{\bysame}{\leavevmode\hbox to3em{\hrulefill}\thinspace}
\providecommand{\MR}{\relax\ifhmode\unskip\space\fi MR }
\providecommand{\MRhref}[2]{%
  \href{http://www.ams.org/mathscinet-getitem?mr=#1}{#2}
}
\providecommand{\href}[2]{#2}

\end{document}